\documentclass[lettersize,journal]{IEEEtran}
\usepackage{amsmath,amsfonts}
\usepackage{algorithmic}
\usepackage{algorithm}
\usepackage{array}
\usepackage[caption=false,font=normalsize,labelfont=sf,textfont=sf]{subfig}
\usepackage{textcomp}
\usepackage{stfloats}
\usepackage{url}
\usepackage{verbatim}
\usepackage{graphicx}
\usepackage{cite}
\usepackage{booktabs}
\usepackage{threeparttable}
\usepackage{makecell}
\usepackage{color}
\usepackage{xcolor}
\usepackage{multicol}
\usepackage{multirow}
\usepackage{soul}
\usepackage{hyperref}
\hypersetup{hidelinks,
	colorlinks=true,
	allcolors=black,
	pdfstartview=Fit,
	breaklinks=true}
\soulregister{\cite}7

\begin{document}

\title{Joint Optimization-based Targetless Extrinsic Calibration for Multiple LiDARs and GNSS-Aided INS of Ground Vehicles}

\author{Junhui Wang\textsuperscript{1,2}, Yan Qiao\textsuperscript{1$\dag$}, Chao Gao\textsuperscript{2$\dag$}, and Naiqi Wu\textsuperscript{1}

\thanks{$^{1}$Institute of Systems Engineering and Collaborative Laboratory for Intelligent Science and Systems, Macau University of Science and Technology, $^{2}$Institute for AI Industry Research (AIR), Tsinghua University.}
}



\maketitle

\begin{abstract}
Accurate extrinsic calibration between multiple LiDAR sensors and a GNSS-aided inertial navigation system (GINS) is essential for achieving reliable sensor fusion in intelligent mining environments. Such calibration enables vehicle–road collaboration by aligning perception data from vehicle-mounted sensors to a unified global reference frame. However, existing methods often depend on artificial targets, overlapping fields of view, or precise trajectory estimation, which are assumptions that may not hold in practice. Moreover, the planar motion of mining vehicles leads to observability issues that degrade calibration performance. This paper presents a targetless extrinsic calibration method that aligns multiple onboard LiDAR sensors to the GINS coordinate system without requiring overlapping sensor views or external targets. The proposed approach introduces an observation model based on the known installation height of the GINS unit to constrain unobservable calibration parameters under planar motion. A joint optimization framework is developed to refine both the extrinsic parameters and GINS trajectory by integrating multiple constraints derived from geometric correspondences and motion consistency. The proposed method is applicable to heterogeneous LiDAR configurations, including both mechanical and solid-state sensors. Extensive experiments on simulated and real-world datasets demonstrate the accuracy, robustness, and practical applicability of the approach under diverse sensor setups.
\end{abstract}


\section{Introduction}
\IEEEPARstart{T}{he} development of intelligent mining systems has significantly advanced the automation and digitalization of open-pit mining operations \cite{Chen2024}. In these complex environments, achieving high levels of safety, efficiency, and autonomy depends critically on the integration of vehicle-mounted and infrastructure-based sensing systems \cite{Xiong2025}. A fundamental requirement in this context is vehicle–road collaboration, where information is shared and fused across multiple platforms to enable accurate perception, localization, and decision-making \cite{Zhang2025,Zhang2024, Annunziata2025,Xiong2025}. To facilitate such collaboration, sensor data acquired from both vehicles and infrastructure must be expressed within a common spatial reference frame, which is typically established using a GNSS-aided inertial navigation system (GINS).

Intelligent mining vehicles are often equipped with multiple LiDAR sensors installed at different locations to extend perception coverage and eliminate blind spots. However, due to variations in sensor types, positions, and orientations, precise extrinsic calibration with respect to the GINS coordinate frame is essential. Without accurate calibration, sensor data cannot be effectively fused, compromising the reliability of perception and localization subsystems.

Existing calibration methods face significant challenges when applied to real-world mining scenarios. Many rely on artificial targets or assume overlapping fields of view between LiDAR sensors \cite{Yan2022,Li2025,Pi2024}. These assumptions are often violated in practice, given the large size of mining vehicles and the complexity of unstructured environments. Furthermore, many methods experience a decline in calibration accuracy when the vehicle undergoes planar motion \cite{Kim2024}. Such motion leads to observability issues, particularly in estimating vertical translation components, which are crucial for aligning sensor data to the global reference frame.

This paper introduces a targetless extrinsic calibration method designed to overcome the practical challenges of sensor alignment in intelligent mining systems. The proposed method calibrates multiple onboard LiDAR sensors with respect to the GINS coordinate system, thereby enabling a unified spatial representation across all vehicle-mounted sensors. It does not require overlapping sensor views, and it is applicable to a wide variety of LiDAR types, including both mechanical spinning and solid-state models.

To improve calibration robustness in the presence of planar motion, an observation model based on the installation height of the GINS unit is introduced. This model provides constraints that help resolve the unobservable components of the calibration problem. Furthermore, a joint optimization framework is developed, which integrates multiple types of constraints derived from sensor measurements to simultaneously refine both the extrinsic parameters and the GINS trajectory.
The main contributions of this work are summarized as follows.

\begin{itemize}
    \item A generalized targetless calibration framework is proposed for aligning multiple LiDAR sensors to a common GINS frame.
    \item A novel installation height-based observation model is introduced to address the observability limitations caused by planar vehicle motion, improving calibration accuracy.
    \item A joint optimization algorithm is developed, incorporating geometric correspondences and motion consistency constraints to refine both sensor extrinsic parameters and GINS trajectory.
    \item The proposed method is validated through extensive experiments using both simulated and real-world datasets, demonstrating its effectiveness under diverse sensor configurations and environmental conditions. To benefit the community, we make our code accessible to the public on \href{https://github.com/KEEPSLAMDUNK/ML-GINS-Calib}{https://github.com/KEEPSLAMDUNK/ML-GINS-Calib}.
\end{itemize}

\section{Related Works}
\subsection{LiDAR-GINS Extrinsic Calibration}
Studies in surveying and mapping related to the calibration of LiDAR-GINS extrinsic parameters primarily involve two categories of methods: approximate and rigorous ones \cite{pentek2020flexible}. Approximate methods aim to reduce discrepancies in georeferenced LiDAR data without incorporating GINS measurements, though they may not address all biases \cite{burman2000calibration,morin2002calibration, habib2010alternative}. Conversely, rigorous methods account for systematic calibration errors and integrate LiDAR and GINS data to correct misalignments \cite{friess2006toward,skaloud2006rigorous, hebel2011simultaneous, kilian1996capture, Underwood2007}. These methods refine calibration by identifying discrepancies through tie points or matching surfaces on overlapping LiDAR strips. The least squares adjustment is often employed to estimate calibration parameters, taking into account additional systematic errors such as laser beam encoder offers, scale factors, and biases in the GINS observations \cite{glira2016rigorous, keyetieu2019automatic}.

Significant efforts have been made in LiDAR-GINS calibration for surveying and mapping. Also, the application of this technology in autonomous driving has been explored. The hand-eye calibration method \cite{horaud1995hand} has been used for calibrating LiDAR and GINS, but it typically offers low estimation accuracy and its results are often used as initial values for other methods. LiDAR-align \cite{Zachary2019} is an open-source method utilizing a nonlinear solver to minimize the distance between each point in the stitching point cloud map and its nearest neighbor. Yan et al. \cite{yan2023extrinsic, Pi2024} propose a method based on bundle adjustment (BA) for extrinsic calibration between LiDAR and GINS, including a mechanism for correcting z-axis offsets using fiducial points. However, its application is limited due to its reliance on a planar feature extraction process and its applicability solely to rotating mechanical LiDARs. Additionally, it needs to address inaccuracies in GINS data caused by vehicle vibrations, which reduce the accuracy of extrinsic parameter estimation. To overcome these limitations, this work jointly optimizes the extrinsic parameters and GINS data.

\subsection{Multi-LiDAR Extrinsic Calibration}
Methods for multi-LiDAR extrinsic calibration can be broadly categorized into target-based and targetless-based methods. Target-based methods focus on determining spatial offsets by integrating data from multiple sensors, necessitating identifiable objects and correspondence points. For example, Gao et al. \cite{gao2010online} use retro-reflective landmarks to calibrate dual LiDAR systems, while Pusztai et al. \cite{pusztai2017accurate} enhance low-resolution LiDAR calibration using corner points from calibration boxes. Xie et al. \cite{xie2018infrastructure} present a calibration framework for multi-camera and multi-LiDAR system using AprilTags to create a suitable calibration environment for autonomous driving platforms. Additionally, Choi et al. \cite{choi2015extrinsic} propose an method to estimate the relative pose between two 2-D LiDARs by scanning orthogonal planes. Jiao et al. \cite{jiao2019novel} develop a method for automatically calibrating multiple LiDARs using three linearly independent planar surfaces. However, target-based methods often require specific environmental conditions and may not be suitable for online calibration.

Targetless-based methods can be further divided into motion-based and appearance-based methods. Motion-based methods calculate trajectories in local coordinates to construct motion constraints for estimating extrinsic parameters. Huang et al. \cite{huang2017extrinsic} introduce an extrinsic multisensor calibration method based on Gauss-Helmert estimation, improving accuracy and robustness by considering pose observation errors. Della Corte et al. \cite{della2019unified} present a unified calibration pipeline for mobile robots with multiple sensors, allowing for concurrent estimation of kinematic, sensor parameters, and time delays. Chang et al. \cite{chang2023versatile} propose an adaptive multi-LiDAR calibration method utilizing global pose graph optimization to estimate extrinsic parameters without requiring overlapping Fields of View. Additionally, Das et al. \cite{das2023observability} leverage GNSS and LiDAR pose matching for multi-LiDAR calibration and propose observability criteria for pose selection. Despite their advantages, these methods are susceptible to inaccuracies in motion estimation and require diverse motion patterns to ensure parameter observability.

Targetless appearance-based methods utilize environmental object appearances to facilitate sensor calibration. Jiao et al. \cite{jiao2019automatic} propose an automatic calibration method that refines calibration parameters by minimizing point-plane distances in multi-LiDAR configurations on autonomous vehicles. Wei et al. \cite{wei2022croon} introduce a two-stage automatic calibration method for multiple LiDARs in road scenarios, leveraging natural road scene features to enhance calibration accuracy. However, these methods typically require overlapping FoVs among LiDARs. To address this, Liu et al. \cite{liu2021extrinsic} propose a targetless calibration method that achieves overlapping FoVs through vehicular movement and utilizes factor graph optimization for the extrinsic calibration of multiple solid-state LiDARs. Liu et al. \cite{liu2022targetless} further improve this method by incorporating adaptive voxelization and LiDAR BA to enhance speed and accuracy. Similarly, the proposed method in this work can also be applied to LiDARs with non-overlapping FoVs and is versatile enough for both solid-state and mechanical LiDARs in structured and unstructured environments.

The method proposed in this work is a targetless appearance-based approach that is capable of jointly calibrating LiDAR-GINS extrinsic parameters and multi-LiDAR extrinsic parameters. It eliminates the need for laser odometry estimation for each LiDAR, a requirement in most other methods. Moreover, this method is suitable for calibrating both mechanical and solid-state LiDARs across diverse environments, \textit{i.e.}, structured or unstructured ones.

\begin{figure*}[thbp]
  \centering
  \includegraphics[scale=0.7]{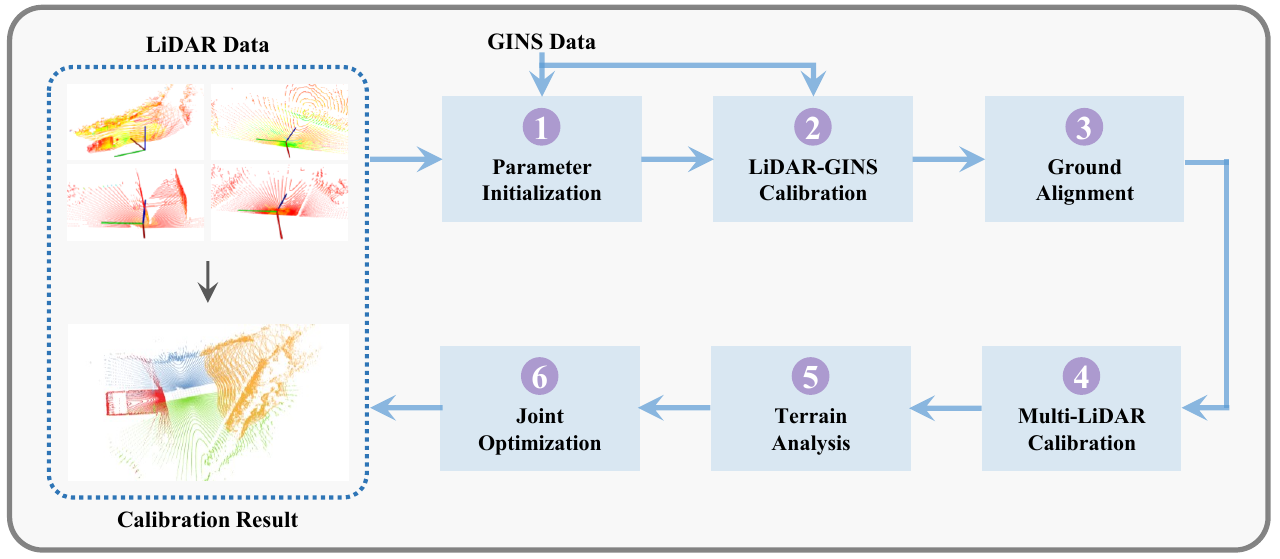}
  \caption{Overview of the calibration workflow. LiDAR-GINS and multi-LiDAR extrinsic parameters are calibrated jointly.}
  \label{system_overview}
\end{figure*}

\section{Preliminary}
\subsection{Notation}
\begin{figure}[t]
  \centering
  \includegraphics[scale=1.0]{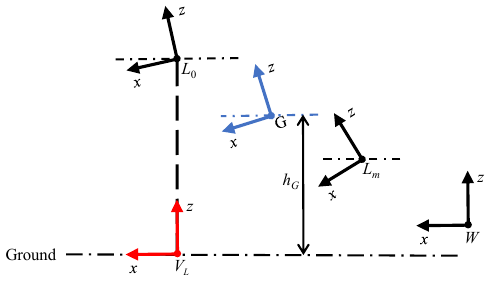}
  \caption{Illustration of the coordinate system used in this work.}
  \label{coordinate_system}
\end{figure}

The illustration of coordinate system in this paper is shown in Fig. \ref{coordinate_system}. The world coordinate system is denoted by $\{W\}$, while the GINS coordinate system is represented by $\{G\}$. The LiDAR coordinate system is represented by $\{L_{m}\}$. The custom virtual LiDAR (VLiDAR) coordinate system is denoted by $\{V_L\}$, with its xy-plane coinciding with the ground. 

Let $^B_A\mathbf{T} \in SE(3)$ be the coordinate transformation from frame A to frame B. The set of $m+1$ LiDARs is denoted by $\mathbb{L}=\{L_0, L_1, \ldots, L_{m}\}$, where $L_0$ represents the base LiDAR. $^{L_0}_{L_{m}}\mathbf{T}$ represents the extrinsic parameters between LiDAR $L_0$ and LiDAR $L_m$, while $^{G}_{L_m}\mathbf{T}$ represents the extrinsic parameters between LiDAR $L_m$ and the GINS system. $^{G}_{V_L}\mathbf{T}$ represents the extrinsic parameters between the GINS system and the virtual LiDAR coordinate system.

Let $\mathbb{T}_{L_m} = \{{^{W}_{L_m}\mathbf{T}_{t_0}}, {^{W}_{L_m}\mathbf{T}_{t_1}}, \dots, {^{W}_{L_m}\mathbf{T}_{t_i}}\}$ be the poses of the LiDAR $L_m$ at time $t_i$ in the world coordinate system, and $\mathbb{T}_{G} = \{{^W_{G}\mathbf{T}_{t_0}}, {^W_{G}\mathbf{T}_{t_1}}, \cdot, {^W_{G}\mathbf{T}_{t_j}}\}$ represents the pose data measured by the GINS system at time $t_j$. A point cloud scanned by LiDAR $L_m$ at time $t_k$ is denoted by ${^{L_m}\mathbb{P}_{t_k}}$.

\subsection{Observation Model of GINS Installation Height}
To address the reduced performance in calibrating LiDAR and GINS systems caused by the planar motion restriction in autonomous vehicles, an observation model is proposed as shown in Eq.\eqref{eq3-B-1}. This observation model requires measuring the GINS equipment installation height relative to the ground. 
\begin{equation} \label{eq3-B-1}
  \begin{aligned}
    \mathbf{y} &= {^{V_L}_{G}\mathbf{T}} \, \mathrm{Exp}(\xi)
  \end{aligned}  
\end{equation}
where $\mathbf{y}=\mathrm{Exp}\left([0, 0, 0, 0, 0, h_G]^\top \right)$ represents the measurement for the GINS installation height. Here, $h_G$ denotes the height difference between the GINS system and the ground. The term $\xi \sim \mathcal{N}\left(0, \mathbf{Q}\right)$ denotes Gaussian noise, with $\mathbf{Q}$ as the covirance matrix defined in Eq. \eqref{eq3-B-2}.
\begin{equation} \label{eq3-B-2}
        {\mathbf{Q}} =
        \left [\begin{matrix}
    c_b & 0 & 0 & 0 & 0 & 0 \\
    0 & c_b & 0 & 0 & 0 & 0 \\
    0 & 0 & c_b & 0 & 0 & 0 \\
    0 & 0 & 0 & c_b & 0 & 0 \\
    0 & 0 & 0 & 0 & c_b & 0 \\
    0 & 0 & 0 & 0 & 0 & c_s \\
    \end{matrix} \right ]
\end{equation}
where $c_b=10^8$ indicates significant uncertainty in one dimension, while $c_s=10^{-3}$ represents small uncertainty. The observation model can be transformed as shown in \eqref{eq3-B-3} for nonlinear optimization, allowing it to be incorporated into the cost function for calibrating extrinsic parameters.
\begin{equation} \label{eq3-B-3}
  \begin{aligned}
    \left\| \mathrm{Log} \left(\mathbf{y} \,  {^{V_L}_G}\mathbf{T}{^{-1}}\right) \right\|_{\mathbf{Q}}^2
  \end{aligned}  
\end{equation}
where Log(·) directly maps the vector space $\mathbb{R}^6$ from $SE(3)$.

\subsection{Generalized Matching Factor} \label{sec:3.2}
This work adopts the generalized matching factor \cite{segal2009generalized}, which is formulated as shown in Eq. \eqref{eq3-C-1}.
\begin{equation} \label{eq3-C-1}
  \begin{aligned}
    \left\| \mathbf{d} \right\|_{\Sigma}^{2} = \mathbf{d}^{\top} \Sigma^{-1} \mathbf{d}
  \end{aligned}  
\end{equation}
where $\mathbf{d}$ represents the difference between two corresponding points, and $\Sigma$ is a covariance matrix that describes the influence of the sample distribution.

The inverse covariance matrix, $\Sigma^{-1}$, can be chosen flexibly. For instance, if $\Sigma^{-1}$ is set to be an identity matrix, the generalized matching factor corresponds to the point-to-point factor. Similarly, the point-to-plane factor can be achieved by setting a corresponding projection matrix. In this work, the covariance matrix is computed based on the covariance matrices of points. The forms of these covariance matrices are detailed in Sec. \ref{sec:4}.

\subsection{Data Collection and Preprocessing}
During the data collection process, all GINS data is continously recorded. In contrast, keyframes of LiDAR data are captured when the vehicle travels beyond certain thresholds, such as a distance of 2 meters or a turn of 30 degrees. The selection of keyframes helps reduce computational costs and increases the excitation of motion between frames. All sensors, including GINS and multiple LiDARs, are synchronized through hardware timestamping to ensure accurate temporal alignment.

\section{Methodology}
\label{sec:4}
\subsection{Overview}
In this work, we propose a method for jointly calibrating LiDAR-GINS extrinsic parameters and multi-LiDAR extrinsic parameters. The overall pipeline of the proposed method is illustrated in Fig. \ref{system_overview}. Initially, LiDAR and GINS data are input into the parameter initialization module, which estimates the initial LiDAR-GINS extrinsic parameters using a global optimization method. Subsequently, the extrinsic parameters between LiDAR and GINS system are further refined through batch optimization. To compensate for the degeneration of LiDAR-GINS extrinsic parameters caused by the planar motion of vehicles, we introduce a ground alignment module. Following this, multi-LiDAR extrinsic parameters are estimated with base LiDAR poses. The terrain analysis module identifies the smoothest local ground to accurately estimate the extrinsic parameters between the base LiDAR and the virtual LiDAR. Finally, all extrinsic parameters are refined through joint optimization.

This workflow ensures valid initial parameter estimation and subsequent refinement, thereby enhancing the overall calibration accuracy of the system. The proposed method not only addresses the challenges posed by planar motion but also identifies the optimal ground fitting through terrain analysis, providing a robust solution for autonomous driving in complex environments.

\begin{figure}[t]
  \centering
  \includegraphics[scale=0.93]{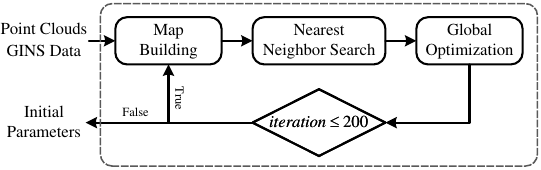}
  \caption{Pipeline for parameter initialization.}
  \label{param_init}
\end{figure}

\subsection{Parameter Initialization}\label{sec:4.2}
The parameter initialization module is responsible for obtaining initial values for LiDAR-GINS extrinsic parameters. Generally, initial values can be acquired either from a computer-aided design (CAD) model or through a hand-eye calibration method \cite{horaud1995hand}. However, in some cases, obtaining initial values may not be feasible due to the unavailability of CAD models or the challenges in accurately estimating trajectory. To address this, we adopt the initialization method from LiDAR-align \cite{Zachary2019}, which conducts a global search in the parameter space to find suitable initial values.

The pipeline of the parameter initialization is illustrated in Fig. \ref{param_init}. Firstly, a map is constructed using LiDAR data and GINS data. Then, for each point in this map, a k-dimensional tree (K-D Tree) is used to search for its nearest-neighboring point. The distances between all pairs of corresponding points are summed to calculate the error. Initial LiDAR-GINS extrinsic parameters can be determined by minimizing the error in Eq. \eqref{eq4-B-2}.
\begin{equation} \label{eq4-B-2}
    \begin{aligned}
        \arg \min\limits_{^{G}_{L_{m}}\mathbf{T}} \sum\limits_{i}{\left\| \mathbf{d}_i\right\|^2}
    \end{aligned}
\end{equation}
where $\mathbf{d}_i$ represents the difference of a pair of corresponding points. Subsequently, the DIviding RECTangles (DIRECT) algorithm is used to solve Eq. \eqref{eq4-B-2} \cite{Jones_1993}. We iteratively build the map and perform optimization until the specified number of iterations is reached (e.g., 200). It should be noted that only the rotational part is optimized, while the translation part is fixed to zero. The restriction is imposed to reduce search space for parameters.

\subsection{LiDAR-GINS Calibration} \label{sec:4.3}
The calibration process for LiDAR and the GINS system is modeled as a batch optimization problem. Initially, distances between pairs of points in two point clouds are computed using Eq. \eqref{eq4-C-1}. The association of corresponding points is based on the K-D tree. 
\begin{equation} \label{eq4-C-1}
    \begin{aligned}
        {^{L_m}\mathbf{d}^{kl}_{t_i t_j}} &= {^{W}_{G}\mathbf{T}_{t_i}} {^{G}_{L_{m}}\mathbf{T}} {\left(^{L_m}\mathbf{p}^k\right)_{t_i}} - {^{W}_{G}\mathbf{T}_{t_j}} {^{G}_{L_{m}}\mathbf{T}} {\left(^{L_m}\mathbf{p}^l\right)_{t_j}}
    \end{aligned}
\end{equation}
where ${\left(^{L_m}\mathbf{p}^k\right)_{t_i}} \in {^{L_m}\mathbb{P}_{t_i}}$ denotes the $k$-th point sampled by $L_m$ at time $t_i$; ${\left(^{L_m}\mathbf{p}^l\right)_{t_j}} \in {^{L_m}\mathbb{P}_{t_j}}$ denotes the $l$-th point sampled by $L_m$ at time $t_j$; ${^{L_m}\mathbf{d}^{kl}_{t_i t_j}}$ represents the difference between these two points after coordinate transformation. The corresponding covariance matrices can be calculated using Eq. \eqref{eq4-C-2}.
\begin{equation} \label{eq4-C-2}
    \begin{aligned}
        \Sigma_{t_it_j}^{kl} &= {^{W}_{G}\mathbf{T}_{t_i}} {^{G}_{L_{m}}\mathbf{T}} {\left(\Sigma^k\right)_{t_i}} ({^{W}_{G}\mathbf{T}_{t_i}} {^{G}_{L_{m}}\mathbf{T}})^{\top} \\
        &+ {^{W}_{G}\mathbf{T}_{t_j}} {^{G}_{L_{m}}\mathbf{T}} {\left(\Sigma^l\right)_{t_j}} ({^{W}_{G}\mathbf{T}_{t_j}} {^{G}_{L_{m}}\mathbf{T}})^{\top}
    \end{aligned}
\end{equation}
where ${\left(\Sigma^k\right)_{t_i}}$ denotes the covariance matrix of ${\left(^{L_m}\mathbf{p}^k\right)_{t_i}}$; ${\left(\Sigma^l\right)_{t_j}}$ denotes the covariance matrix of ${\left(^{L_m}\mathbf{p}^l\right)_{t_j}}$; $\Sigma_{t_it_j}^{kl}$ represents the combined covariance matrix used in the optimization process. The covariance matrix of each point is estimated from its $k$ neighbors (e.g., $k=20$).

Next, LiDAR-GINS extrinsic parameters can be estimated using Eq. \eqref{eq4-C-3}. In practice, the Huber robust kernel function is usually used to reduce the impact of erroneous data associations.
\begin{equation} \label{eq4-C-3}
    \begin{aligned}
        arg \min\limits_{{^{G}_{L_{m}}\mathbf{T}}} \sum\limits_{L_m} \sum\limits_{t_i,t_j} \sum\limits_{k,l} \rho \left( \left\| {^{L_m}\mathbf{d}^{kl}_{t_i t_j}} \right\| ^2_{\Sigma_{t_it_j}^{kl}} \right)
    \end{aligned}
\end{equation}
where $\rho(\cdot)$ denotes the Huber kernel function.

It should be noted that uneven terrain, common in environments such as open-pit mines and wilderness areas, can induce vibrations in a vehicle's body. These vibrations result in deviations within the GINS data, potentially leading to error when estimating the extrinsic parameters. Therefore, in the joint optimization stage, GINS poses is considered as variables and is optimized concurrently with LiDAR-GINS extrinsic parameters to yield a more precise estimation.

\begin{figure}[t]
  \centering
  \includegraphics[scale=1.1]{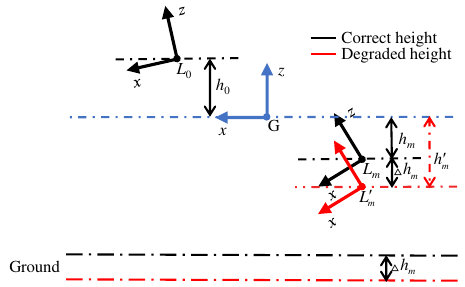}
  \caption{Illustration of the unobservable case in LiDAR-GINS calibration.}
  \label{ground_alignment}
\end{figure}

\subsection{Ground Alignment} \label{sec:4.4}
The ground alignment module aims to compensate for degraded LiDAR-GINS extrinsic parameters. The unobservable phenomenon is illustrated in Fig. \ref{ground_alignment}. In the coordinate system $\{G\}$, the xy-plane is parallel to the ground. When LiDAR-GINS extrinsic parameters are fully observable, the relative poses of coordinate systems are correct, as shown in black in Fig. \ref{ground_alignment}. In addition, the ground detected by different LiDARs should overlap.

However, when a vehicle trajectory follows a nearly planar path, it lacks significant motion perpendicular to the ground. In this situation, the relative height component in LiDAR-GINS extrinsic parameters becomes unobservable \cite{lv2022observability}. Therefore, compensating for these extrinsic parameters is necessary. One approach to address this issue is by aligning the ground observed by other LiDARs with the ground observed by the base LiDAR.

To begin with, a point cloud is randomly selected for each LiDAR and transformed to the coordinate system $\{G\}$ using LiDAR-GINS extrinsic parameters estimated in Sec. \ref{sec:4.3}. Following this, ground equations are extracted from this point cloud. The ground equation can be expressed as Eq. \eqref{eq4-D-0}.
\begin{equation} \label{eq4-D-0}
        n_1x + n_2y + n_3z + d = 0
\end{equation}
where $\mathbf{n} = [n_1, n_2, n_3]$ represents the normal vector to the ground, and $d$ denotes the intercept of the ground plane. In the end, the relative heights are calculated by using Eq. \eqref{eq4-D-1}.
\begin{equation} \label{eq4-D-1}
        \triangle h_m = d_m - d_0
\end{equation}
where $d_m$ and $d_0$ represent the intercepts of ground equations, while $\triangle h_m$ denotes the relative height used for compensation. Degraded LiDAR-GINS extrinsic parameters can be compensated for using Eq. \eqref{eq4-D-2}.
\begin{equation} \label{eq4-D-2} 
    {^G_{L_m}\mathbf{T}}^{*} =
    \begin{bmatrix}
    1 & 0 & 0 & 0 \\
    0 & 1 & 0 & 0 \\
    0 & 0 & 1 & \triangle h_m \\
    0 & 0 & 0 & 1
    \end{bmatrix}
    \cdot
     {^G_{L_m}\mathbf{T}}
\end{equation}
where ${^G_{L_m}\mathbf{T}}^{*}$ denotes LiDAR-GINS extrinsic parameters after compensation.

\subsection{Multi-LiDAR Calibration}
The extrinsic parameters of multiple LiDARs and the base LiDAR poses are estimated jointly. Initially, the initial values for base LiDAR poses are computed using LiDAR-GINS extrinsic parameters and GINS data, as shown in Eq. \eqref{eq4-E-1}.
\begin{equation} \label{eq4-E-1}
    \begin{aligned}
        \mathbb{T}_{L_0}^{*} = \Big\{ &{^{W}_{L_0}\mathbf{T}_{t_i}^{*}} = {^{W}_{G}\mathbf{T}_{t_i}} {^{G}_{L_0}\mathbf{T}^{*}} \Big\}
    \end{aligned}
\end{equation}
where $\mathbb{T}_{L_0}^{*}$ denotes the set of initial values for poses of base LiDAR $L_0$ at time $t_i$. Similarly, the distances between any two points in any two point clouds are calculated by Eq. \eqref{eq4-E-2}. It should be noted that the association of corresponding points is based on the K-D tree.
\begin{equation} \label{eq4-E-2}
    \begin{aligned}
        \left( \mathbf{d}^{kl}_{t_i t_j} \right)_{mn} &= {^{W}_{L_0}\mathbf{T}_{t_i}} {^{L_0}_{L_m}\mathbf{T}}{\left(^{L_m}\mathbf{p}^k\right)_{t_i}} - {^{W}_{L_0}\mathbf{T}_{t_j}} {^{L_0}_{L_n}\mathbf{T}} {\left(^{L_n}\mathbf{p}^l\right)_{t_j}}
    \end{aligned}
\end{equation}
where $\left( \mathbf{d}^{kl}_{t_i t_j} \right)_{mn}$ represents the difference between the $k$-th point sampled by LiDAR $L_m$ at $t_i$ and the $l$-th point sampled by LiDAR $L_n$ at $t_j$. It should be noted that $m$ and $n$ can be the same. The corresponding covariance matrices can be calculated by Eq. \eqref{eq4-E-3}.
\begin{equation} \label{eq4-E-3}
    \begin{aligned}
        \left( \Sigma^{kl}_{t_i t_j} \right)_{mn} &= {^{W}_{L_0}\mathbf{T}_{t_i}} {^{L_0}_{L_m}\mathbf{T}}{\left(^{L_m}\Sigma^k\right)_{t_i}} ({^{W}_{L_0}\mathbf{T}_{t_i}} {^{L_0}_{L_m}\mathbf{T}})^{\top} \\
        &+ {^{W}_{L_0}\mathbf{T}_{t_j}} {^{L_0}_{L_n}\mathbf{T}}{\left(^{L_n}\Sigma^l\right)_{t_j}} ({^{W}_{L_0}\mathbf{T}_{t_j}} {^{L_0}_{L_n}\mathbf{T}})^{\top}
    \end{aligned}
\end{equation}
where ${\left(^{L_m}\Sigma^k\right)_{t_i}}$ denotes the covariance matrix of ${\left(^{L_m}\mathbf{p}^k\right)_{t_i}}$; ${\left(^{L_n}\Sigma^l\right)_{t_j}}$ denotes the covariance matrix of ${\left(^{L_n}\mathbf{p}^l\right)_{t_j}}$; $\left( \Sigma^{kl}_{t_i t_j} \right)_{mn}$ represents the combined covariance matrix to be used in the optimization process. The covariance matrix of each point is estimated from its $k$ neighbors. Subsequently, the base LiDAR poses and multi-LiDAR extrinsic parameters can be estimated by Eq. \eqref{eq4-E-4}.
\begin{equation} \label{eq4-E-4}
    \begin{aligned}
        \arg \min\limits_{\mathbb{T}_{L_0},{^{L_0}_{L_m}\mathbf{T}}} \sum\limits_{m,n} \sum\limits_{t_i,t_j} \sum\limits_{k,l} \rho \left( \left\| \left( \mathbf{d}^{kl}_{t_i t_j} \right)_{mn} \right\| ^2_{\left(\Sigma_{t_it_j}^{kl}\right)_{mn}} \right)
    \end{aligned}
\end{equation}
where $\mathbb{T}_{L_0}$ denotes the set of base LiDAR poses to be estimated; ${^{L_0}_{L_m}\mathbf{T}}$ represents the extrinsic parameters between $L_0$ and $L_m$.

\begin{figure}[t]
  \centering
  \includegraphics[scale=1.1]{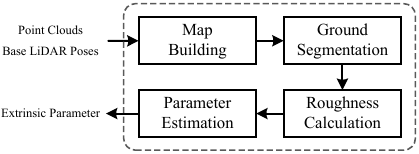}
  \caption{Process of terrain analysis aimed at finding the most accurate extrinsic parameter between the base LiDAR and virtual LiDAR.}
  \label{terrain_analysis}
\end{figure}

\subsection{Terrain Analysis} \label{sec:4.5}
To improve the accuracy of extrinsic parameter estimation between the base LiDAR and the virtual LiDAR, it is crucial to account for the variability in terrain. In environments where the ground is not uniformly flat, the precision of the parameter estimation may be inaccurate. To minimize the impact of uneven ground on the estimation process, we have implemented a terrain analysis method. This method involves selecting the flattest ground segment on which the extrinsic parameter estimation is performed. The process of this method are illustrated in Fig \ref{terrain_analysis}.

To begin with, the point cloud map is built based on the base LiDAR poses estimated in Sec. \ref{sec:4.4}. Next, the patchwork++ mthod is used to extract the ground \cite{Lee2022}. Afterward, the local ground is cropped for each base LiDAR pose. The roughness of each local ground is then computed using Eq. \eqref{eq4-F-1}. 
\begin{equation} \label{eq4-F-1}
    \begin{aligned}
        r = \sqrt{\frac{1}{N} \sum\limits^{N}_{i=1}\left(d_i - \overline{d} \right)}
    \end{aligned}
\end{equation}
where $r$ represents the roughness; $d_i$ denotes the distance of the $i$-th point to the fitted ground plane model, and $\overline{d}$ denotes the average distance.

Finally, the local ground with the lowest roughness is used to estimate the extrinsic parameter between the base LiDAR and virtual LiDAR using Eq. \eqref{eq4-F-2} and Eq. \eqref{eq4-F-3}.

\begin{equation} \label{eq4-F-2}
    \begin{cases}
        \mathbf{u} = \frac{\mathbf{n}_1 \times \mathbf{n}_2}{\left\| \mathbf{n}_1 \times \mathbf{n}_2 \right\|}\\
        \theta = \arccos{(\mathbf{n}_1\cdot\mathbf{n}_2)}\\
        \mathbf{R} = \mathbf{I}+[\mathbf{u}]_{\times}\sin{\theta}+[\mathbf{u}]_{\times}^2(1-\cos{\theta})\\
        \mathbf{t} = [0, 0, d_1-d_2]^{\mathrm{T}}\\
    \end{cases}
\end{equation}
\begin{equation} \label{eq4-F-3}
        {_{L_0}^{V_L}\mathbf{T}} = 
        \left [\begin{matrix}
    \mathbf{R} & \mathbf{t} \\
    0 & 1 
    \end{matrix} \right ]
\end{equation}
where $[\mathbf{n}_1^{\mathrm{T}},d_1]$ define the parameterized representations of the local ground, $[\mathbf{n}_2^{\mathrm{T}},d_2] = [0, 0, 1, 0]^{\mathrm{T}}$ define the ground plane in the virtual LiDAR coordinate system, $\mathbf{u}$ stands for a unit vector orthogonal to both $\mathbf{n}_1$ and $\mathbf{n}_2$, $\theta$ denotes the angle between $\mathbf{n}_1$ and $\mathbf{n}_2$, $\mathbf{R}$ is derived through Rodrigues' rotation formula, and $\mathbf{t}$ represents the translation component.

\subsection{Joint Optimization}
In the joint optimization step, LiDAR data and GINS data are tightly coupled to estimate refined extrinsic parameters. This process considers three types of factors.
\begin{enumerate}
    \item GINS Installation Height Factor: This factor imposes a constraint on the unobservable direction of the VLiDAR-GINS extrinsic parameter, accounting for the plane motion of vehicles. It is mathematically expressed as Eq. \eqref{eq4-G-1}.
    \item General Matching Factors 1 and 2: These involve multi-LiDAR and VLiDAR-GINS extrinsic parameters, respectively. They minimize the difference between points, as described by Eq. \eqref{eq4-G-2} and Eq. \eqref{eq4-G-3}. It should be noted that the GINS poses are also treated as optimized variables, though they are considered as measurements in Eq. \eqref{eq4-C-1}. The reason for this is illustrated in Sec. \ref{sec:4.3}.
    \item Motion Constraint Factor: This factor encompasses the base LiDAR poses, GINS poses, and VLiDAR-GINS extrinsic parameters, detailed in Eq. \eqref{eq4-G-4}.
\end{enumerate}

\begin{equation} \label{eq4-G-1}
  \begin{aligned}
    {^{1}\mathbf{r}} = \mathrm{Log} \left(\mathbf{y} \,  {^{V_L}_G}\mathbf{T}{^{-1}}\right)
  \end{aligned}  
\end{equation}
\begin{equation} \label{eq4-G-2}
    \begin{aligned}
        {^{2}\mathbf{r}^{kl}_{t_i t_j}} &= {^{W}_{G}\mathbf{T}_{t_i}} {^{G}_{V_L}\mathbf{T}} {^{V_L}_{L_{0}}\mathbf{T}} {\left(^{L_0}\mathbf{p}^k\right)_{t_i}} - {^{W}_{G}\mathbf{T}_{t_j}} {^{G}_{V_L}\mathbf{T}} {^{V_L}_{L_{0}}\mathbf{T}} {\left(^{L_0}\mathbf{p}^k\right)_{t_j}}
    \end{aligned}
\end{equation}
\begin{equation} \label{eq4-G-3}
    \begin{aligned}
        {^{3}\mathbf{r}^{kl}_{t_i t_j}} &= {^{W}_{L_0}\mathbf{T}_{t_i}} {^{L_0}_{L_m}\mathbf{T}}{\left(^{L_m}\mathbf{p}^k\right)_{t_i}} - {^{W}_{L_0}\mathbf{T}_{t_j}} {^{L_0}_{L_n}\mathbf{T}} {\left(^{L_n}\mathbf{p}^l\right)_{t_j}}
    \end{aligned}
\end{equation}
\begin{equation} \label{eq4-G-4}
    \begin{aligned}
         {^{4}\mathbf{r}_{t_i t_j}} = \mathrm{Log} \left({^{V_L}_{G}}\mathbf{T} \,
        {^{W}_{G}}\mathbf{T}_{t_j}^{-1} \, {^{W}_{G}}\mathbf{T}_{t_i} \,  {^{V_L}_{G}}\mathbf{T}^{-1} \,   {^{W}_{L_0}\mathbf{T}_{t_i}^{-1}} \, {^{W}_{L_0}\mathbf{T}_{t_j}} \right)
    \end{aligned}
\end{equation}
These three types of factors are integrated into a unified cost function, shown in Eq. \eqref{eq4-G-5}, which is then optimized using the Levenberg-Marquardt (LM) optimizer.
\begin{equation} \label{eq4-G-5}
    \begin{aligned}
         f(\hat{\mathcal{X}}) = &\left\| {^{1}\mathbf{r}} \right\|^2_{^{1}\Sigma} + \sum\limits_{t_i,t_j} \sum\limits_{k,l} \left\| {^{2}\mathbf{r}^{kl}_{t_i t_j}} \right\|^2_{^{2}\Sigma} + \\ &\sum\limits_{t_i,t_j} \sum\limits_{k,l} \left\| {^{3}\mathbf{r}^{kl}_{t_i t_j}} \right\|^2_{^{3}\Sigma} + \sum\limits_{t_i,t_j} \left\| {^{4}\mathbf{r}_{t_i t_j}} \right\|^2_{^{4}\Sigma}
    \end{aligned}
\end{equation}
where ${^{1}\Sigma}$, ${^{2}\Sigma}$, ${^{3}\Sigma}$ and ${^{4}\Sigma}$ represent the corresponding covariance matrices of ${^{1}\mathbf{r}}$, ${^{2}\mathbf{r}^{kl}_{t_i t_j}}$, ${^{3}\mathbf{r}^{kl}_{t_i t_j}}$ and ${^{4}\mathbf{r}_{t_i t_j}}$, respectively.

The optimization process yields the VLiDAR-GINS and multi-LiDAR extrinsic parameters. Then, the extrinsic parameters between the base LiDAR and the GINS system can be calculated using Eq. \eqref{eq4-G-6}.
\begin{equation} \label{eq4-G-6}
    \begin{aligned}
         {^{L_0}_{G}}\mathbf{T} = {^{L_0}_{V_L}\mathbf{T}} \, {^{V_L}_{G}}\mathbf{T}
    \end{aligned}
\end{equation}

\begin{figure}[t]
  \centering
  \includegraphics[scale=0.55]{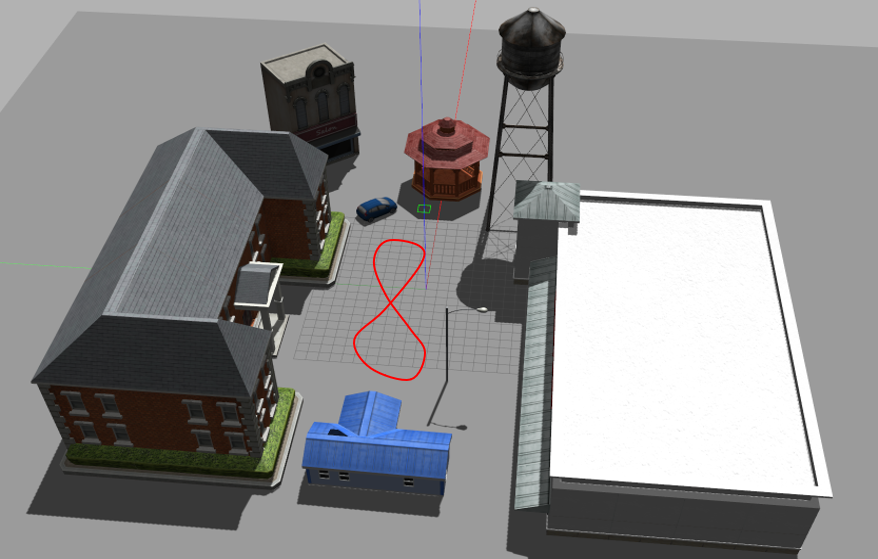}
  \caption{The simulated environment used to generate the dataset. The car follows the red 8-character trajectory to collect data.}
  \label{sim_env}
\end{figure}

After each optimization step, motion compensation is applied to the point clouds based on the optimized poses. This ensures higher accuracy and consistency in subsequent iterations.

\section{Experiments} \label{sec:5-B}
To our knowledge, no existing open-source methods fully align with our goals. Thus, we compare the LiDAR-GINS extrinsic parameters and multi-LiDAR extrinsic parameters obtained through our method with those from related works.

\subsection{Experiment Setup}
In this work, both simulated and real-world datasets are utilized to validate the proposed method.
\subsubsection{Simulated Dataset}
The constructed simulation environment is illustrated in Fig. \ref{sim_env}. In this setup, the vehicle collects data following an 8-character trajectory. The simulated dataset includes data from a Livox Avia LiDAR (solid-state), a 16-channel Velodyne LiDAR (rotating), and GINS. The Velodyne LiDAR is designated as the base LiDAR. Notably, the two LiDARs do not have overlapping fields of view, which presents a calibration challenge and underscores the method's ability to handle such scenarios.

In the simulated environment, the ground truth of the extrinsic parameters can be directly obtained. This provides a straightforward benchmark for evaluating the accuracy of the proposed calibration method.
\subsubsection{Real-world Dataset}
\begin{figure}[t]
  \centering
  \includegraphics[scale=0.65]{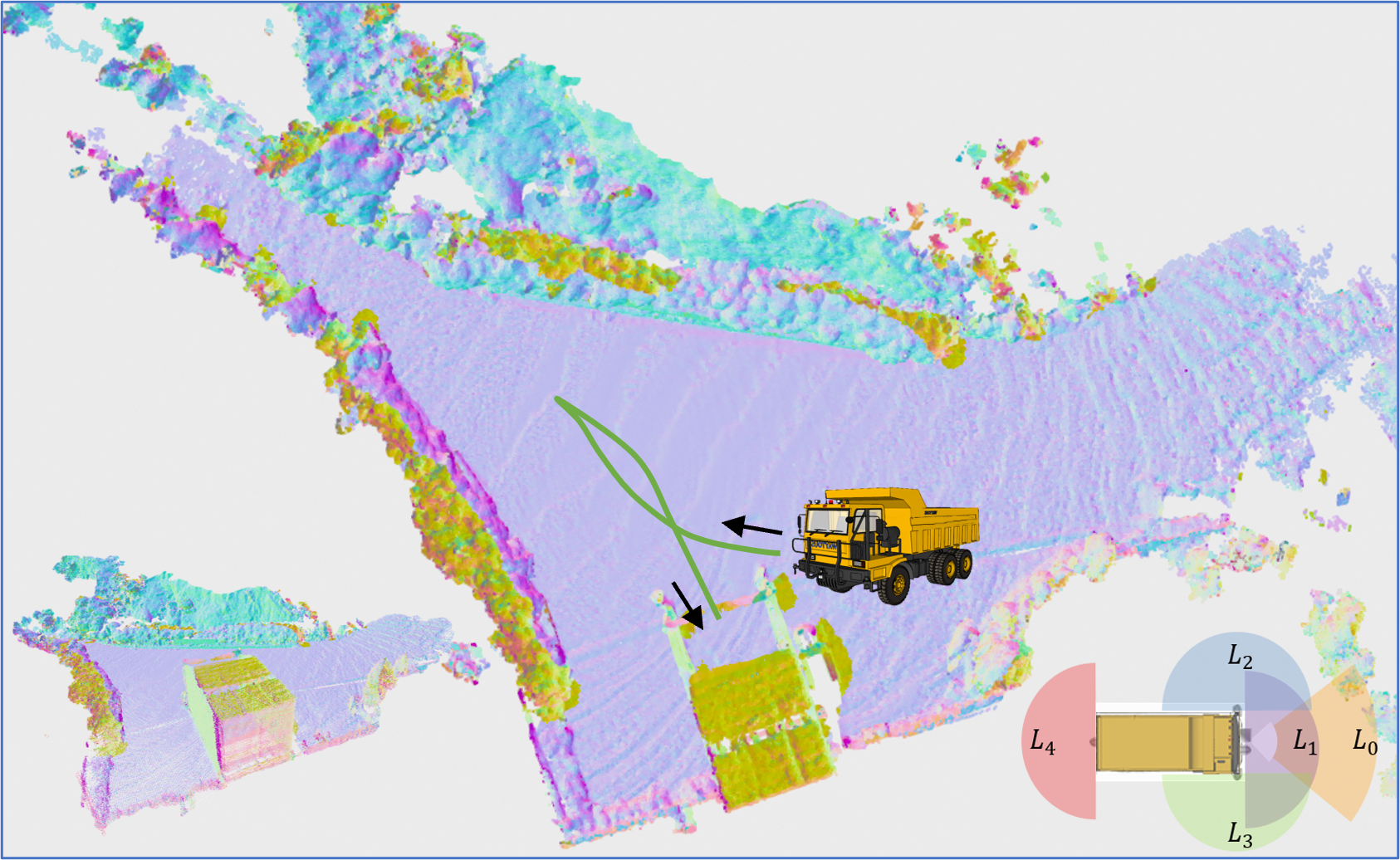}
  \caption{The data is collected by a truck equipped with five LiDARs and a GINS in an open-pit mine. $L_0$ is a Livox Horizon, and $L_1 \sim L_4$ are Robosense Bpearl LiDARs. The truck follows the green curve trajectory to collect data. The background features a reconstructed model colored according to surface normals.}
  \label{trajectory}
\end{figure}

For the real-world dataset, data was collected using a truck platform equipped with five LiDARs and a GINS, as despicted in Fig. \ref{trajectory}. The GINS system used provides a position accuracy of approximately 3 cm. The base LiDAR, $L_0$, is a Livox Horizon LiDAR (solid-state), featuring an $81.7^{\circ}$ horizontal FoV and a range accuracy of 2 cm. The other LiDARs, $L_1 \sim L_4$, are Robosense Bpearl LiDARs (rotating) with a $360^{\circ}$ FoV and a range accuracy of 1 cm. These additional LiDARs are placed to monitor blind spots. All LiDARs are synchronized at 10 Hz using hardware. The real-world data was gathered from an unstructured open-pit mine, providing a challenging and realistic environment for testing the proposed method. The driving trajectory of the truck, which is used to collect this data, is illustrated in Fig. \ref{trajectory}.

Obtaining ground truth for LiDAR-GINS extrinsic parameters in the real world presents significant challenges. As a result, we evaluate the accuracy of these parameters by examining map consistency, as detailed in Section \ref{sec:5-B}.

To establish a reference solution for evaluating multi-LiDAR extrinsic parameters, all LiDAR sensors mounted on the truck are registered to a 64-channel LiDAR, which serves as a reliable basis for comparison. The accuracy and reliability of this reference solution are rigorously verified to ensure the credibility of the evaluation.
\subsubsection{Experimental Platform}
The proposed algorithm has been implemented on a high-performance PC featuring an Intel i9-13980HK processor.

\begin{table*}[thbp]
\renewcommand\arraystretch{1.5}
\centering
\caption{Evaluation of LiDAR-GINS Extrinsic Parameters for Real-world Dataset Using MME and MPV Metrics}
\label{lidar-gins-eva}
\begin{tabular}{@{}ccccccccccc@{}}
\toprule
               & \multicolumn{2}{c}{$L_0$} & \multicolumn{2}{c}{$L_1$} & \multicolumn{2}{c}{$L_2$} & \multicolumn{2}{c}{$L_3$} & \multicolumn{2}{c}{$L_4$} \\
               & MME{$\ \downarrow$}        & MPV{$\ \downarrow$}       & MME{$\ \downarrow$}        & MPV{$\ \downarrow$}       & MME{$\ \downarrow$}        & MPV{$\ \downarrow$}       & MME{$\ \downarrow$}        & MPV{$\ \downarrow$}       & MME{$\ \downarrow$}        & MPV{$\ \downarrow$}       \\ \midrule
HECalib \cite{Daniilidis1999}  & -2.81774          & 0.236816         & -          & -         & -          & -         & -          & -         & -          & -         \\
LiDAR-align \cite{Zachary2019}    & --2.80735          & 0.236921         & -3.2655          & 0.20153         & -2.92135          & 0.229419         & -3.32815          & 0.207739         & -3.22371          & 0.189405         \\
Ours & \textbf{-3.24945}          & \textbf{0.182268}         & \textbf{-4.26837}          & \textbf{0.11642}         & \textbf{-4.02182}          & \textbf{0.125006}         & \textbf{-3.9693}          & \textbf{0.148375}         & \textbf{-3.73137}          & \textbf{0.163125}         \\ \bottomrule
\end{tabular}
\begin{tablenotes}
    \item - denotes the failure of estimation.
\end{tablenotes}
\end{table*}

\subsection{Experimental Evaluation Metrics}
To ensure the accuracy of LiDAR-GINS and multi-LiDAR extrinsic parameters, we employ three evaluation metrics.
\subsubsection{Mean Map Entropy and Mean Plane Variance}
According to \cite{razlaw2015evaluation}, we use mean map entropy (MME)  and mean plane variance (MPV) to assess map consistency. Lower MME and MPV values indicate better map consistency, which reflects higher accuracy in extrinsic parameters. MME quantifies the uncertainty in the map by measuring the entropy of the point cloud. It is calculated as follows.
\begin{equation} \label{eq16}
    \begin{aligned}
        \text{MME} = \frac{1}{n} \sum^{n}_{k=1}\frac{1}{2} \ln \left | 2 \pi e \Sigma (q_k) \right|
    \end{aligned}
\end{equation}
where $\Sigma(q_k)$ is the sample covariance of point $q_k$ within a local radius $r$ (e.g., 1.0m), and $n$ is the total number of points in the map.
MPV measures the flatness of the map by approximating a plane from 3D points within a given radius and calculating the variance of the distances of points to this plane. It is defined as follows.
\begin{equation} \label{eq17}
    \begin{aligned}
        \text{MPV} = \frac{1}{n} \sum^{n}_{k=1} v(q_k)
    \end{aligned}
\end{equation}
where $v(q_k)$ is the upper quartile of distances within the specified radius (e.g., 1.0m), and $n$ is the number of points in the map.
\subsubsection{Distance for Lie Groups}
When ground truth is available, we calculate the difference between the estimated extrinsic parameters and the ground truth using the following metrics.
\begin{equation} \label{eq18}
    \begin{cases}
        R_{err} =  \left\| \mathrm{Log}\left({\mathbf{R}^{-1}_{gt}}\mathbf{R}_{ext} \right) \right\|^{2} \\
         \  t_{err} =  \left \| {\mathbf{t}_{gt}} - \mathbf{t}_{ext} \right\|^{2}
    \end{cases}
\end{equation}
where ${\mathbf{R}_{gt}}$ and ${\mathbf{t}_{gt}}$ are the rotation and translation parts of the ground truth of multi-LiDAR extrinsic parameters, and ${\mathbf{R}_{ext}}$ and ${\mathbf{t}_{ext}}$ are the rotation and translation parts of the estimated extrinsic parameters.

For the simulated dataset, we directly compute the difference between the estimated extrinsic parameters and the ground truth using Eq. \eqref{eq18}. For the real-world dataset, we assess the accuracy of the stitched maps created using GINS and LiDAR data. Higher quality stitched maps indicate more accurate LiDAR-GINS extrinsic parameters. The accuracy of multi-LiDAR extrinsic parameters is evaluated using the same metric describe in Eq. \eqref{eq18}.

\begin{table}[t]
\renewcommand\arraystretch{1.5}
\centering
\caption{Evaluation of Extrinsic Parameters for Simulated Dataset}
\label{sim-eva}
\begin{tabular}{@{}ccccccccccc@{}}
\toprule
        & \multicolumn{2}{c}{${^{G}_{L_0}\mathbf{T}}$} & \multicolumn{2}{c}{${^{L_0}_{L_1}\mathbf{T}}$}\\
        & $R_{err} (\deg)$      & $t_{err} (\text{m})$     & $R_{err} (\deg)$       & $t_{err} (\text{m})$     \\ \midrule
HECalib \cite{Daniilidis1999} & 0.084    & 85.644  & 155.671  & 216.121          \\
LiDAR-align \cite{Zachary2019}   & 0.233  & 1.244  & 0.427   & 1.117   \\
Ours*   & 0.115  & 0.337  & \textbf{0.066}   & \textbf{0.01}   \\
Ours    & \textbf{0.056}  & \textbf{0.031}  & \textbf{0.066}   & \textbf{0.01}   \\ \bottomrule
\end{tabular}
\begin{tablenotes}
    \item Ours* denotes the proposed method but without terrain analysis.
\end{tablenotes}
\end{table}

\subsection{Calibration Results of Simulated Dataset}
We compare the proposed method with HECalib and LiDAR-align. For HECalib, the laser odometry for the two LiDARs is estimated using DLO \cite{Chen2022}, followed by the computation of the extrinsic parameters using hand-eye calibration. For LiDAR-align, it estimates the LiDAR-GINS extrinsic parameters for the two LiDARs, which are then used to determine the extrinsic parameters between the two LiDARs. Additionally, we introduce noise into the estimated extrinsic parameters between the base LiDAR and the VLiDAR to simulate the impact of not using terrain analysis. This result is labeled as Ours*. All results are presented in Table \ref{sim-eva}.

The translation component of $^{L_0}_{L_1}T$ in both HECalib and LiDAR-align suffers from severe degeneration due to planar motion when evaluating LiDAR-GINS extrinsic parameters. The proposed method, utilizing an observation model, effectively addresses this issue. However, its performance decreases without terrain analysis. Regarding multi-LiDAR extrinsic parameters, HECalib yields almost entirely incorrect results due to the inaccurate laser odometry estimation for Avia in the simulated environment.

In conclusion, the proposed method achieves the highest accuracy across all metrics. HECalib's accuracy depends on the precision of laser odometry, which can be unstable across different scenarios. While LiDAR-align shows stable performance, it suffers from degeneration in unobservable directions. Terrain analysis is beneficial for LiDAR-GINS calibration in the proposed method.

\begin{figure}[t]
  \centering
  \includegraphics[scale=0.42]{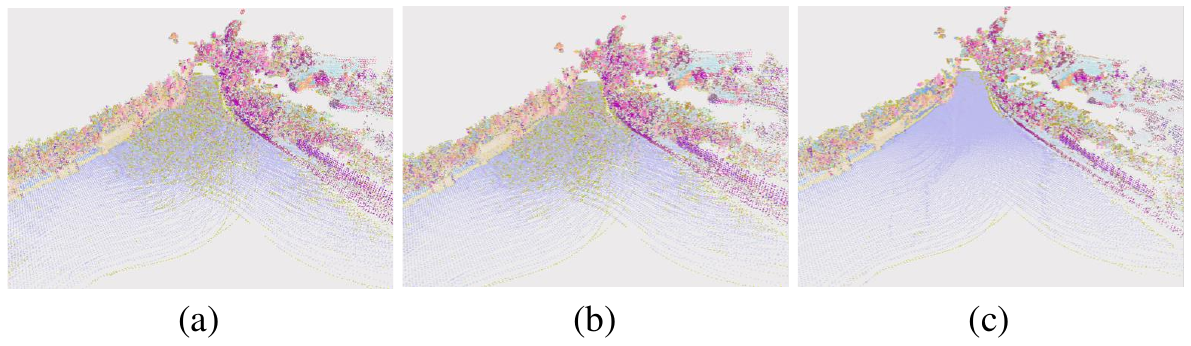}
  \caption{Maps generated using base LiDAR data and the LiDAR-GINS extrinsic parameter. (a) and (b) show the calibration results of HECalib and LiDAR-align, respectively. (c) illustrates the calibration result of the proposed method. Maps are colored based on the normal of each point, reflecting the smoothness of object surfaces. Map quality is evaluated based on color consistency.}
  \label{lidar_gins}
\end{figure}

\subsection{Calibration Results of Real-world Dataset}
\subsubsection{LiDAR-GINS Calibration Results}
The proposed method is evaluated against HECalib \cite{Daniilidis1999} and LiDAR-align \cite{Zachary2019}. To estimate trajectory for all LiDARs, the DLO \cite{Chen2022} method is utilized. However, it can only correctly estimate the base LiDAR trajectory. Consequently, HECalib is restricted to calibrating only the base LiDAR and the GINS system. Thus, the calibration results are demonstrated through maps stitched from the base LiDAR data, as shown in Fig. \ref{lidar_gins}. The maps are color-coded based on the normal vector of each point, which reflects the smoothness of the object surfaces. It is clear that the ground color consistency in Fig. \ref{lidar_gins}(c) surpasses that in Fig. \ref{lidar_gins}(a) and Fig. \ref{lidar_gins}(b), indicating that the proposed method achieves higher accuracy.

Table \ref{lidar-gins-eva} provides a quantitative assessment using MME and MPV metrics. HECalib method achieves accuracy in $L_0$ comparable to that of LiDAR-algin. This demonstrates that HECalib's accuracy heavily relies on trajectory accuracy, which limits its application in varied environments. In contrast, the LiDAR-align method effectively calibrates all LiDAR-GINS extrinsic parameters with good accuracy. The proposed method surpasses both HECalib and LiDAR-align, highlighting the benefit of jointly optimizing GINS poses.

The results demonstrate that the proposed method considerably improves the calibration accuracy of LiDAR-GINS extrinsic parameters, resulting in improved map consistency and overall performance. The joint optimization approach, which refines GINS poses, proves more effective than HECalib and LiDAR-align.

\begin{table*}[t]
\renewcommand\arraystretch{1.5}
\centering
\caption{Evaluation of Multi-LiDAR Extrinsic Parameters for Real-world Dataset}
\label{multi-lidar-eva}
\begin{tabular}{@{}ccccccccccc@{}}
\toprule
        & \multicolumn{2}{c}{${^{L_0}_{L_1}\mathbf{T}}$} & \multicolumn{2}{c}{${^{L_0}_{L_2}\mathbf{T}}$} & \multicolumn{2}{c}{${^{L_0}_{L_3}\mathbf{T}}$} & \multicolumn{2}{c}{${^{L_0}_{L_4}\mathbf{T}}$} & \multicolumn{2}{c}{RMSE} \\
        & $R_{err} (\deg)$      & $t_{err} (\text{m})$     & $R_{err} (\deg)$       & $t_{err} (\text{m})$    & $R_{err} (\deg)$       & $t_{err} (\text{m})$    & $R_{err} (\deg)$       & $t_{err} (\text{m})$    & $R_{err} (\deg)$        & $t_{err} (\text{m})$     \\ \midrule
VGICP \cite{Koide2021}   & 0.165  & 0.053  & 0.465   & 0.065  & 0.524   & 0.315  & 0.614   & 0.355  & 0.473    & 0.241   \\
GICP \cite{segal2009generalized}    & 0.161  & 0.027  & \textbf{0.158}   & \textbf{0.022}   & 0.443   & 0.078   & 0.488   & 0.215    & 0.348    & 0.116   \\
NDT \cite{Biber2003}     & 0.186  & 0.090  & 0.306   & 0.112  & 0.359   & 0.143  & 1.471    & 0.448  & 0.778    & 0.246   \\
Ours    & \textbf{0.122}  & \textbf{0.018}  & 0.348   & 0.037  & \textbf{0.171}   & \textbf{0.051}  & \textbf{0.422}   & \textbf{0.149}  & \textbf{0.293}    & \textbf{0.081}   \\ \bottomrule
\end{tabular}
\end{table*}

\begin{table*}[t]
\renewcommand\arraystretch{1.5}
\centering
\caption{The Running Time of Calibration}
\label{running_time}
\begin{tabular}{@{}cccccccc@{}}
\toprule
\makebox[0.1\textwidth][c]{Datasets}   & \makebox[0.1\textwidth][c]{\begin{tabular}[c]{@{}c@{}}The number\\ of Scans\end{tabular}} & \makebox[0.1\textwidth][c]{\begin{tabular}[c]{@{}c@{}}LiDAR-GINS\\ Calibration\end{tabular}} & \makebox[0.1\textwidth][c]{\begin{tabular}[c]{@{}c@{}}Ground\\ Alignment\end{tabular}} & \makebox[0.1\textwidth][c]{\begin{tabular}[c]{@{}c@{}}Multi-LiDAR\\ Calibration\end{tabular}} & \makebox[0.1\textwidth][c]{\begin{tabular}[c]{@{}c@{}}Terrain\\ Analysis\end{tabular}} & \makebox[0.1\textwidth][c]{\begin{tabular}[c]{@{}c@{}}Joint\\ Optimization\end{tabular}} & \makebox[0.1\textwidth][c]{\begin{tabular}[c]{@{}c@{}}Total\end{tabular}} \\ \midrule
Simulated  & 29x2         & 1.759s             & 0.007s                                     & 4.216s                  & 0.922s          & 5.375s     &     12.279s                                                   \\
Real-world & 19x5       & 16.546s       & 0.09s          & 25.832s                      & 2.512s         & 103.185s     & 148.165s                                              \\ \bottomrule
\end{tabular}
\end{table*}

\subsubsection{Multi-LiDAR Calibration Results}
The proposed method is evaluated against HECalib \cite{Daniilidis1999}, VGICP \cite{Koide2021}, GICP \cite{segal2009generalized} and NDT \cite{Biber2003}. We manually selected point clouds from the dataset that are beneficial for registration, which allows VGICP, GICP, and NDT to achieve better results. This process was not applied to the proposed method in this work. As shown in Table \ref{multi-lidar-eva}, HECalib fails to estimate extrinsic parameters due to poor trajectory estimation for LiDAR $L_1$ to $L_4$. GICP shows superior performance in estimating ${^{L_0}_{L_2}\mathbf{T}}$, while the proposed method achieves the best results in ${^{L_0}_{L_1}\mathbf{T}}$, ${^{L_0}_{L_3}\mathbf{T}}$ and ${^{L_0}_{L_4}\mathbf{T}}$.
The Root Mean Square Error (RMSE) indicates that the proposed method achieves the highest accuracy overall, with approximately 0.293 degrees in rotation and 0.081 meters in translation. However, the precision for ${^{L_0}_{L_4}\mathbf{T}}$ is suboptimal across all methods. This is attributed to the limited FoV of the rear LiDAR. 

\begin{figure}[t]
  \centering
  \includegraphics[scale=0.5]{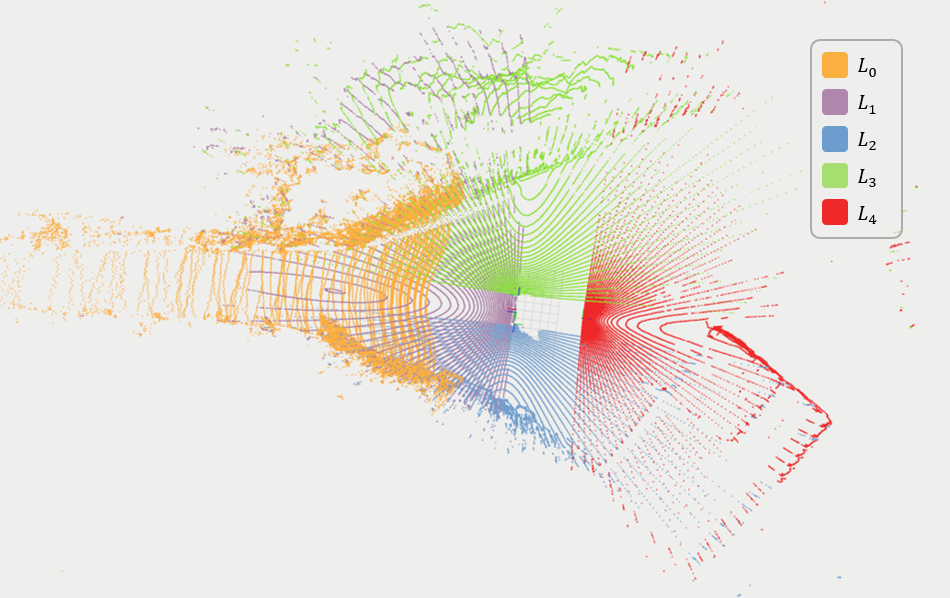}
  \caption{Visualization of the calibration results for multiple LiDARs in the real-world dataset.}
  \label{multi-lidars-align}
\end{figure}

In conclusion, the results demonstrate the effectiveness of the proposed algorithm in achieving precise extrinsic calibration without requiring specific types of LiDAR or environments. The alignment results of all LiDARs are illustrated in Fig. \ref{multi-lidars-align}.

\subsection{Running Time of Calibration}
It is crucial to present the running time of the proposed method, as detailed in Table \ref{running_time}. The optimization process involves 29 LiDAR scans for the simulated environment and 19 LiDAR scans for the real-world environment. As shown, the majority of the time consumption occurs during the LiDAR-GINS calibration, multi-LiDAR calibration, and joint optimization stages.

As the number of sensors and the volume of data increase, the computational workload grows exponentially. However, the total running time remains acceptable for the setups used in this work. We can also reduce the computational workload by employing sliding-window optimization, though this may decrease calibration accuracy. This trade-off depends on the real-time requirements of the specific cases.

\section{Conclusions}
This work introduces a versatile, targetless extrinsic calibration method that simultaneously calibrates the extrinsic parameters between LiDAR and GINS, as well as between multiple LiDARs. Our method eliminates the need for overlapping fields of view and demonstrates strong adaptability across various LiDAR types and environments. To accommodate vehicles in planar motion, an observation model is proposed to constrain the unobservable components in LiDAR-GINS calibration. Additionally, a joint optimization approach based on three types of factors is presented to further enhance calibration accuracy. Extensive experiments with both simulated and real-world datasets validate the efficacy of our method, demonstrating its applicability to both mechanical and solid-state LiDARs and confirming its high calibration accuracy. 

Future work will focus on extending the proposed calibration framework to incorporate camera sensors.

\bibliographystyle{IEEEtran}
\bibliography{IEEEabrv,ref}

\end{document}